\documentclass{article}

\usepackage[final]{neurips_2024}

\usepackage[utf8]{inputenc} 
\usepackage[T1]{fontenc}    
\usepackage{hyperref}       
\usepackage{url}            
\usepackage{booktabs}       
\usepackage{amsfonts}       
\usepackage{nicefrac}       
\usepackage{microtype}      
\usepackage{xcolor}         
\usepackage{colortbl}       
\usepackage{array}          

\definecolor{tblheader}{HTML}{2E4A6B} 
\definecolor{tblstripe}{HTML}{EEF1F5} 
\newcommand{\headrow}[1]{\textbf{\textcolor{white}{#1}}}
\usepackage{graphicx}
\usepackage{bm}
\usepackage{amsfonts}
\usepackage{amsmath}
\usepackage{caption}
\usepackage{subcaption}
\usepackage{lipsum}

\usepackage{hyperref}

\usepackage[LGRgreek]{mathastext}

\title{MAM-CLIP: Vision--Language Pretraining on Mammography Atlases for BI-RADS Classification}

\author{%
  Halil Ibrahim Gulluk \\
  Department of Electrical Engineering\\
  Stanford University\\
  Stanford, CA 94305 \\
  \texttt{gulluk@stanford.edu} \\
  \And 
  Olivier Gevaert \\
  Biomedical Informatics Research (BMIR)
\\
  Stanford University\\
  Stanford, CA 94304 \\
  \texttt{ogevaert@stanford.edu} \\
}

\begin{document}

\maketitle

\begin{abstract}
    Deep learning methods have demonstrated promising results in predicting BI-RADS scores from mammography images. However, the interpretation of these images can vary, leading to discrepancies even among radiologists. Given the inherent complexity of mammography images, training classification models solely on image labels often yields limited performance. To address this challenge, we curated 2{,}313 mammogram images and their corresponding captions from two mammography atlases. Our proposed approach employs a multi-modal model that uses a pretrained PubMedBERT as the language component. By training this model on image--text pairs with contrastive learning, we enable the vision encoder to absorb the rich information contained in the captions, thereby improving its understanding of mammography findings. We then fine-tune the vision encoder on two datasets for BI-RADS prediction, achieving superior performance compared with models trained without this pretraining, particularly when labeled samples are scarce. The improvement in the 3-class average F1 score ranges from +1\% to +14\%, depending on the number of training samples: a +1\% increase with 40K training samples, and a +14\% increase with 1K samples. Furthermore, our experiments reveal that 2K image--text pairs from mammography atlases can be more informative than 2K labeled samples even for label prediction, with an average margin of +1.1\% when more than 10K training samples are available, underscoring the value of incorporating textual information when modeling medical images. Overall, our work provides a vision-language model for mammography and highlights the value of textual information from mammography atlases.
 In addition, to facilitate future research, we publicly release preprocessed mammography images of the TEKNOFEST dataset \cite{teknofest_source}. The training code, pre-trained model weights, data extraction scripts, and the released dataset are publicly available at: \href{https://github.com/igulluk/MAM-CLIP}{https://github.com/igulluk/MAM-CLIP}
\end{abstract}

\section{Introduction}
Breast cancer is a leading cause of cancer mortality. In 2020, the World Health Organization reported that 2.26 million individuals were diagnosed with breast cancer \cite{WHO_cancer}. Mammography, MRI, and ultrasound are commonly used for both the diagnosis and screening of breast cancer. Mammography, in particular, is widely used because of its rapid image acquisition, which makes it well suited to routine clinical evaluation. This study focuses on developing models based on mammography images.

Mammography involves two primary views: the cranio-caudal (CC) and the mediolateral-oblique (MLO) view. In practice, each patient undergoes four acquisitions: LCC and LMLO for the left breast, and RCC and RMLO for the right breast. Findings in mammography are typically reported using the Breast Imaging Reporting and Data System (BI-RADS) \cite{liberman2002breast}, which serves as the standard tool.

The BI-RADS system categorizes mammography images into seven classes that indicate the risk of breast cancer. The BI-RADS scores and their corresponding risks are detailed in Table \ref{tab:birads}. This study aims to predict BI-RADS scores from mammography images.

In recent years, researchers have focused on developing deep learning models for breast cancer detection and BI-RADS score prediction \cite{shen2019deep,tsai2022high,liu2021deep,nguyen2022novel}. In \cite{tsai2022high}, the authors used segmentation masks of breast lesions to select small patches based on their overlap with the lesions, and trained deep convolutional models on these patches for BI-RADS classification. In \cite{nguyen2022novel}, separate models were trained for BI-RADS and density classification for each view (LCC, LMLO, RCC, and RMLO), after which the LightGBM algorithm \cite{ke2017lightgbm} was used for the final prediction.

\begin{table}[t]
\centering
\caption{BI-RADS scores and their corresponding diagnoses and descriptions.}\label{tab:birads}
\renewcommand{\arraystretch}{1.25}
\setlength{\tabcolsep}{8pt}
\begin{tabular}{@{}lll@{}}
\toprule
\rowcolor{tblheader}
\headrow{BI-RADS Score} & \headrow{Diagnosis} & \headrow{Description} \\
\midrule
BI-RADS 1 & Negative & No finding; normal \\
\rowcolor{tblstripe}
BI-RADS 2 & Benign & Definite benign finding \\
BI-RADS 3 & Probably benign & Findings benign with probability ${>}\,98\%$ \\
\rowcolor{tblstripe}
BI-RADS 0 & Incomplete & Further information needed for diagnosis \\
BI-RADS 4 & Suspicious findings & Possibility of malignancy ($3\%$--$94\%$) \\
\rowcolor{tblstripe}
BI-RADS 5 & Highly suspicious of malignancy & Possibility of malignancy (${>}\,95\%$) \\
BI-RADS 6 & Positive & Biopsy-proven malignancy \\
\bottomrule
\end{tabular}
\end{table}

In a separate study, the authors proposed models for breast cancer detection that leverage electronic health records in addition to mammogram images \cite{akselrod2019predicting}. In \cite{shen2019deep}, deep convolutional models were proposed for both patch-based and whole-image classification of breast cancer from mammograms. The combination of mammography images and clinical factors for estimating the malignancy of microcalcifications was studied in \cite{liu2021deep}.

More broadly, researchers have been developing vision-language models for the medical domain \cite{moor2023med,zhang2023large,wang2022medclip,lin2023pmc,zhang2023pmc}. These models have proven useful either through fine-tuning for specific tasks or through visual question answering. However, these works typically use datasets with very few or no mammography images, and the accompanying captions often lack detailed explanations of the underlying diseases, as they are not written for educational purposes. This underscores the value of our image--text dataset. In our study, we extract images and their captions from radiology atlases. These atlases are explicitly designed to educate radiologists, so their captions are information-rich and carefully explain the details within each image to support accurate diagnosis.

In contrast to classical computer vision datasets, medical images often require more nuanced interpretation. This is particularly true for BI-RADS classification, where clinical interpretation can be decisive. The BI-RADS 0 class, which indicates a need for further information, adds to this complexity: clinicians may assign a BI-RADS 0 label for various reasons that can be strongly associated with other classes. To address this challenge, we developed a multi-modal model that accepts mammogram images together with corresponding captions from mammography atlases. This approach enables our models to capture the complex information in breast images, as well as the reasons behind how and why images are labeled with specific BI-RADS scores. Our results demonstrate a significant improvement in BI-RADS classification through the integration of vision-language information.

\section{Methodology}

\subsection{Vision-Language Model}

Our objective is to train a vision-language model similar to CLIP \cite{radford2021learning}, in which images and their captions are mapped to similar representations, as measured by cosine similarity. For the language component, we use the pretrained PubMedBERT model \cite{gu2021domain}. For the visual component, we use a ConvNeXt model \cite{liu2022convnet} pretrained on ImageNet. We observe that using high-resolution mammography images is crucial for good classification performance. We therefore do not use pretrained transformer-based vision encoders from vision-language models such as Med-Flamingo \cite{moor2023med} and BiomedCLIP \cite{zhang2023large}, since they are trained on low-resolution images. We further find that ResNet-based models \cite{he2016deep} achieve lower accuracy than ConvNeXt models, and consequently we also avoid the ResNet-based vision encoders used in models such as MedCLIP \cite{wang2022medclip}, PMC-CLIP \cite{lin2023pmc}, and PMC-VQA \cite{zhang2023pmc}.

We adopt the training methodology of PMC-CLIP \cite{lin2023pmc}, in which the vision-language model is trained with an InfoNCE loss that increases the similarity of matched image--text pairs, together with a masked language modeling (MLM) loss. The MLM loss encourages the vision encoder to be as predictive as possible of the masked language tokens.

To formalize, let $\Phi_{vis}$ and $\Phi_{text}$ denote the ConvNeXt visual encoder and the PubMedBERT text encoder, respectively, and let the current batch be

$\mathcal{B} = \{(\mathcal{I}_1, \mathcal{T}_1), (\mathcal{I}_2, \mathcal{T}_2), \ldots, (\mathcal{I}_N, \mathcal{T}_N)\}$, where $(\mathcal{I}_i, \mathcal{T}_i)$ denotes the $i$-th image--text pair. Following the convention in \cite{lin2023pmc}, we denote the image representations by $\bm{v}_i=\Phi_{vis}(\mathcal{I}_i)$ and the text representations by $\bm{T}_i=\Phi_{text}(\mathcal{T}_i)$, where $\bm{v}_i \in \mathbb{R}^d$ and $\bm{T}_i \in \mathbb{R}^{l\times d}$. Here $l$ denotes the text length and $d$ the embedding dimension. We further denote the \texttt{[CLS]} token representation by $\bm{t}_i \in \mathbb{R}^d$, and the cosine similarity between $\bm{v}_i$ and $\bm{t}_j$ by $s_{ij}$. The contrastive learning loss is then
\[
\mathcal{L}_{contrastive} = \dfrac{-1}{N}\sum_{i=1}^N \log \bigg (\dfrac{exp(s_{ii})}{\sum_{j=1}^N exp(s_{ij})}\bigg ) - \dfrac{1}{N}\sum_{i=1}^N \log \bigg (\dfrac{exp(s_{ii})}{\sum_{j=1}^N exp(s_{ji})}\bigg )
\]

\vspace{-0.5cm}
\begin{figure}
\centering
\includegraphics[width=\textwidth]{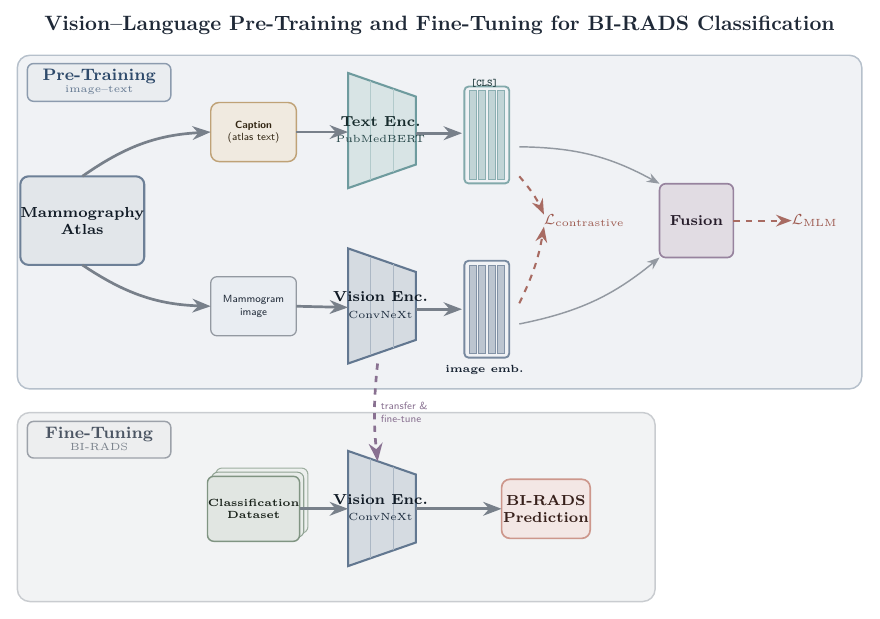}
\caption{Model overview. We first extract mammogram images and their corresponding captions from mammography atlases and train a vision-language model using a contrastive loss and a masked language modeling loss. The text encoder is a pretrained PubMedBERT. We then fine-tune the vision encoder for BI-RADS and density classification on two datasets.} \label{fig:model_overview}
\end{figure}

Following the masked language modeling approach of \cite{lin2023pmc}, we randomly mask words with a fixed probability and predict the masked words using not only the text itself but also the vision embedding from the visual encoder. Specifically, we use a transformer fusion model $\Phi_{fusion}$ that takes the image embedding and the masked text embedding and predicts the ground-truth text sequence. We denote the prediction of the fusion model for the masked token by $\bm{p}_i = \Phi_{fusion}(\bm{v}_i, \bm{T}_{i}^{masked})$. If the ground truth is $\bm{y}_i$, the overall MLM loss is
\[
\mathcal{L}_{MLM} = \mathbb{E}_{\mathcal{B}}[\mathcal{L}_{CE}(\bm{y}_i,\bm{p}_i))]
\]
where $\mathcal{L}_{CE}$ is the cross-entropy loss and the expectation is taken over all masked tokens in the batch. With a weight $\lambda$ for the MLM loss, the overall pretraining loss is
\[
\mathcal{L} = \mathcal{L}_{contrastive} + \lambda\cdot \mathcal{L}_{MLM}
\]

\subsection{Pretraining Dataset Preparation}
We extract mammography images and their corresponding captions from two mammography atlases: the \emph{Atlas of Mammography} \cite{de2007atlas} and the \emph{ACR BI-RADS Atlas} \cite{acr-birads}. For the \emph{Atlas of Mammography}, we used the Python library PyMuPDF \cite{PyMuPDF} to extract the images and captions. For the \emph{ACR BI-RADS Atlas}, we additionally used the pytesseract library \cite{pytesseract} to recover the text from the images via optical character recognition. When a caption describes more than one image, we pair each image with that caption as a separate image--caption pair. The Python scripts used to extract the image--text pairs are available in our code.

\begin{figure}[!t]
\centering
\includegraphics[width=0.75\textwidth]{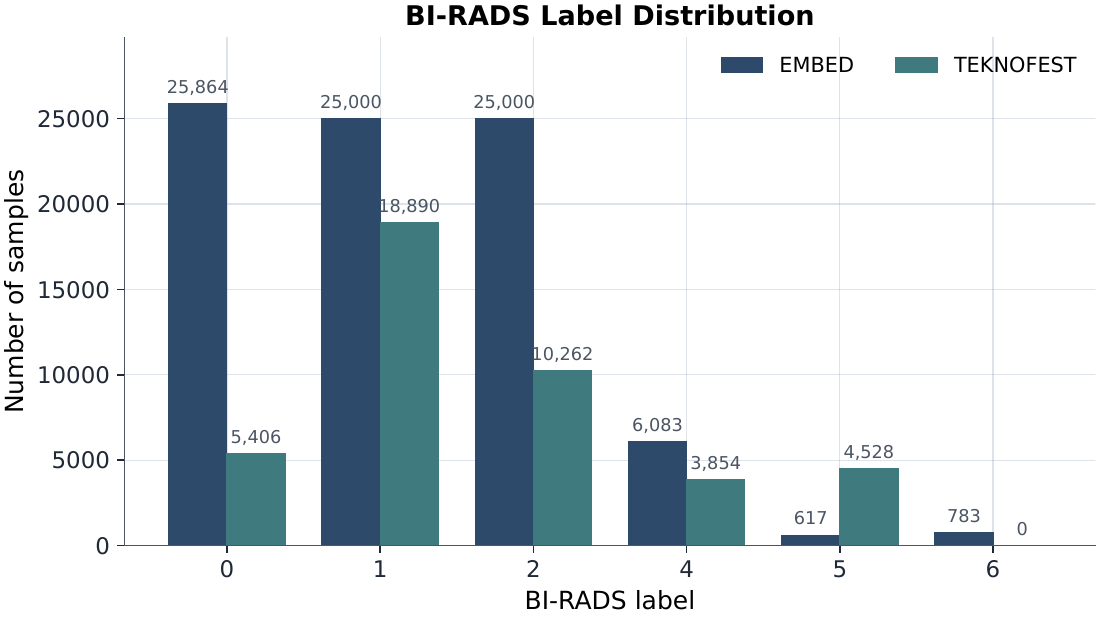}
\caption{Number of samples per class for the two classification datasets. The EMBED dataset is highly imbalanced with respect to the BI-RADS labels, and the TEKNOFEST dataset contains no BI-RADS 6 samples.} \label{fig:distributions}
\end{figure}

\section{Classification Datasets}
Our ultimate goal is to predict the BI-RADS class. To evaluate our models, we use two datasets: the Emory Breast Imaging Dataset (EMBED) \cite{jeong2023emory} and the TEKNOFEST Artificial Intelligence in Healthcare Competition 2023 dataset \cite{Teknofest}.

\textbf{EMBED.} The original dataset contains views other than MLO and CC; we use only the MLO and CC views. The dataset is also highly imbalanced, as approximately $73\%$ of the images have a BI-RADS 1 label. To mitigate this imbalance, we cap the number of images with a BI-RADS 1 or BI-RADS 2 label at $25{,}000$ per class, and we exclude the BI-RADS 3 class from our experiments. The resulting label distribution is shown in Fig. \ref{fig:distributions}.

\textbf{TEKNOFEST.} The TEKNOFEST dataset \citep{saglik2024dataset} was prepared for the 2023 Artificial Intelligence in Health Competition \cite{teknofest2024} by the Republic of Turkey Ministry of Health, and its clinical characteristics were described in the associated publication \cite{teknofest_source}. It comprises data from $10{,}735$ patients, with up to four images per patient (RMLO, RCC, LMLO, and LCC), for a total of $42{,}940$ images; the label distribution is shown in Fig. \ref{fig:distributions}. We release the preprocessed PNG images and an image-level metadata file to facilitate future research; access details are provided in our repository.\footnote{\url{https://github.com/igulluk/MAM-CLIP}} 
DICOM images from both datasets are preprocessed with a YOLOX model \cite{ge2021yolox} to crop the breast region from the background of the original DICOM images. Further details on this preprocessing step are provided in the code.

\section{Experiments}

We first train our multi-modal model on image--text pairs and select the checkpoint with the lowest validation loss. We then fine-tune both the ImageNet-pretrained model and the vision encoder of our vision-language model (VLM) on the two classification datasets. In all experiments, we merge the BI-RADS 6 images into the BI-RADS 5 class and exclude the BI-RADS 3 class, which is present only in the EMBED dataset. We therefore work with five BI-RADS classes, and we use 4-fold cross-validation for all experiments.

\subsection{Implementation Details}
For the visual and text encoders, we initialize an ImageNet-pretrained ConvNeXt \cite{liu2022convnet} and a pretrained PubMedBERT \cite{gu2021domain}, respectively. Unlike many other multi-modal models, we use a relatively high image resolution of $1024\times768$ pixels. We use a batch size of 64 and the AdamW optimizer \cite{loshchilov2017decoupled} with a learning rate of $1\mathrm{e}{-}4$, and we train for 25 epochs. All experiments are run on a single NVIDIA A100 GPU.

\subsection{Results}

For BI-RADS classification, we consider a 5-class task. We additionally examine performance on a simplified 3-class grouping, in which we merge BI-RADS 1 with BI-RADS 2 and BI-RADS 4 with BI-RADS 5, yielding the three classes \{BI-RADS 1,\,2\}, \{BI-RADS 0\}, and \{BI-RADS 4,\,5\}. This grouping is clinically meaningful, as the three classes can be interpreted as benign, requiring further information, and malignant, respectively.

\begin{figure}
     \centering
     \makebox[\linewidth][c]{%
     \begin{subfigure}[b]{0.60\textwidth}
     
         \centering
         \includegraphics[width=\textwidth]{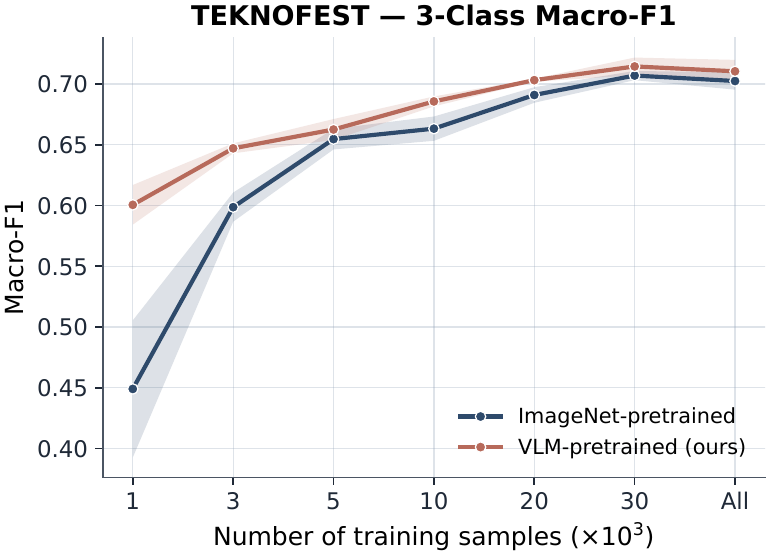}
         \caption{}
         \label{fig:t3_3birads}
     \end{subfigure}
     
     \hfill
     \begin{subfigure}[b]{0.60\textwidth}
         \centering
         \includegraphics[width=\textwidth]{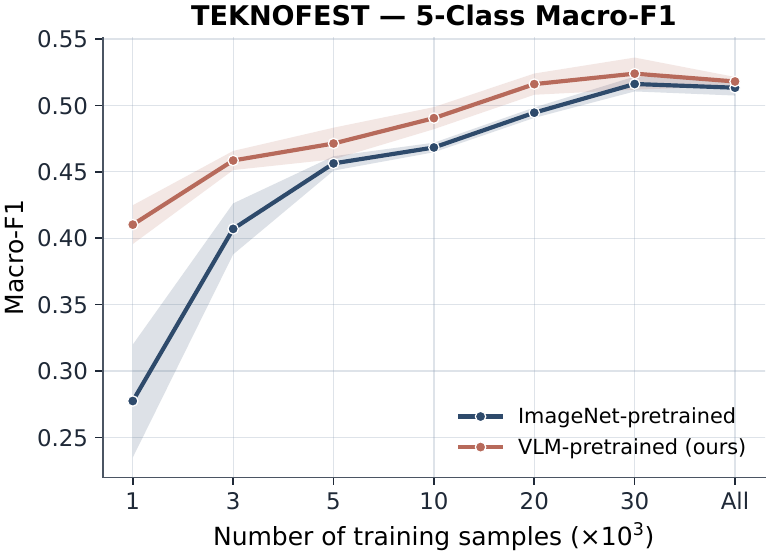}
         \caption{}
         \label{t3_5birads}
     \end{subfigure}%
}\\
        \caption{Classification results on the TEKNOFEST dataset. For every number of training samples $n$ (1, 3, 5, 10, 20, 30, All(44), in thousands), our vision-language pretraining outperforms the ImageNet-pretrained model, and the gap between the two models is largest when few training samples are available. All experiments use 4-fold cross-validation.}
        \label{fig:T3_BIRADS}
\end{figure}

\begin{figure}[h!]
     \centering
     \makebox[\linewidth][c]{%
     \begin{subfigure}[b]{0.60\textwidth}
         \centering
         \includegraphics[width=\textwidth]{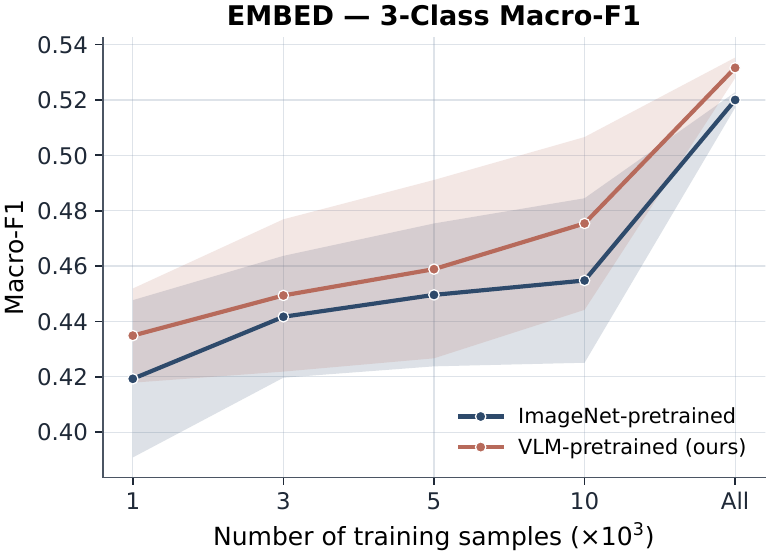}
         \caption{}
         \label{fig:embed_3birads}
     \end{subfigure}
     
     \hfill
     \begin{subfigure}[b]{0.60\textwidth}
         \centering
         \includegraphics[width=\textwidth]{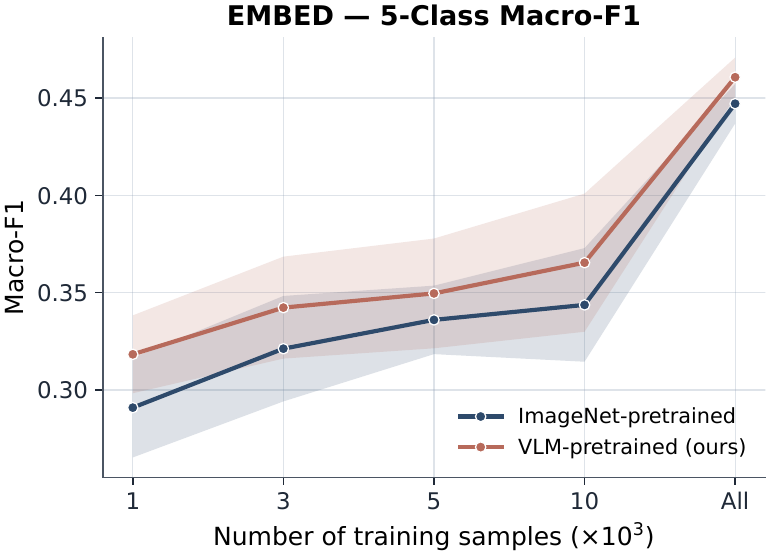}
         \caption{}
         \label{fig:embed_5birads}
     \end{subfigure}%
}\\
        \caption{Classification results on the EMBED dataset. For every number of training samples, our vision-language pretraining outperforms the ImageNet-pretrained model. All experiments use 4-fold cross-validation.}
        \label{fig:EMBED_BIRADS}
\end{figure}

We compute class-wise F1 scores and report the macro-averaged F1 score for both the 3-class and 5-class settings. The results for the TEKNOFEST and EMBED datasets are shown in Fig. \ref{fig:T3_BIRADS} and Fig. \ref{fig:EMBED_BIRADS}, respectively. On both datasets, pretraining on mammogram images and their captions improves overall performance across all numbers of training samples.

\subsection{Assessing Textual Information: Captions vs. Labels}
\begin{table}

\centering
\caption{3-class macro-averaged F1 on the TEKNOFEST dataset. In each row, the ImageNet-pretrained model is given 2{,}000 more training samples than our VLM-pretrained model. For sample sizes larger than 10K, pretraining with 2{,}313 image--text pairs is more beneficial than adding 2{,}000 labeled samples. The better result in each row is shown in bold.}\label{table:birads3_ablation}
\renewcommand{\arraystretch}{1.3}
\setlength{\tabcolsep}{10pt}
\begin{tabular}{@{}cccc@{}}
\toprule
\rowcolor{tblheader}
\headrow{VLM \# Samples} & \headrow{ImageNet \# Samples} & \headrow{VLM Avg.\ F1 (Ours)} & \headrow{ImageNet Avg.\ F1} \\
\midrule
1K  & 3K        & $0.597{\pm}0.015$          & $\mathbf{0.597{\pm}0.012}$ \\
\rowcolor{tblstripe}
3K  & 5K        & $0.646{\pm}0.007$          & $\mathbf{0.654{\pm}0.008}$ \\
10K & 12K       & $\mathbf{0.685{\pm}0.004}$ & $0.668{\pm}0.008$          \\
\rowcolor{tblstripe}
20K & 22K       & $\mathbf{0.701{\pm}0.003}$ & $0.692{\pm}0.007$          \\
30K & 32K       & $\mathbf{0.711{\pm}0.008}$ & $0.697{\pm}0.006$          \\
\rowcolor{tblstripe}
41K & All (43K) & $\mathbf{0.706{\pm}0.002}$ & $0.700{\pm}0.008$          \\
\bottomrule
\end{tabular}
\end{table}

The previous results show that image--text pretraining improves performance. We further demonstrate experimentally that the captions accompanying the pretraining images can provide more informative cues than additional labeled images, even though the downstream task is label prediction. Using our pretraining dataset of $2{,}313$ image--text pairs, we run experiments in which the ImageNet-pretrained model is given $2{,}000$ more samples than our VLM-pretrained model for the classification task. As shown in Table \ref{table:birads3_ablation}, for $n>10{,}000$, adding $2{,}000$ extra labeled samples yields a smaller improvement than pretraining with the original $2{,}313$ pairs, even though the pretraining images come from a completely different source and distribution. This suggests that, owing to the complexity of mammography images, captions from mammography atlases can guide deep models more effectively than additional image labels.

\section{Conclusion}
In conclusion, we have shown that the explanatory captions accompanying mammogram images in mammography atlases can play a crucial role in image classification. Unlike other large pretrained models, using only about $2{,}300$ image--text pairs from mammography atlases leads to significant performance improvements, which highlights the informative nature of these captions. Our experiments further indicate that these captions can be more informative than the image labels themselves. 


\clearpage
\bibliographystyle{unsrt}
\bibliography{refs}

\end{document}